\documentclass[10pt,twocolumn,letterpaper]{article}

\usepackage[pagenumbers]{iccv} 

\definecolor{iccvblue}{rgb}{0.21,0.49,0.74}
\usepackage[pagebackref,breaklinks,colorlinks,allcolors=iccvblue]{hyperref}
\usepackage{amsmath, amssymb, booktabs, subcaption, caption, multirow, makecell}
\usepackage{colortbl}
\usepackage{pgfplots}
\usepackage{pgfplotstable}
\usepackage{tikz}
\usepackage{adjustbox}

\makeatletter
\let\AtPageUpperLeft\@undefined
\let\AtTextLowerLeft\@undefined
\let\AtTextCenter\@undefined
\let\AtPageLowerLeft\@undefined
\let\AtTextUpperLeft\@undefined
\let\AtPageCenter\@undefined
\let\AddToShipoutPicture\@undefined
\let\ClearShipoutPicture\@undefined
\let\ESO@HookI\@undefined
\let\ESO@HookII\@undefined
\let\ESO@HookIII\@undefined
\let\ESO@HookIV\@undefined
\let\ESO@HookV\@undefined
\let\ESO@yoffsetI\@undefined
\let\ESO@yoffsetII\@undefined
\let\ESO@isMEMOIR\relax
\makeatother

\usepackage{pdfpages}

\def\paperID{19} 
\def\confName{ICCV}
\def\confYear{2025}

\title{Edge-case Synthesis for Fisheye Object Detection: A Data-centric Perspective}

\author{
Seunghyeon Kim\thanks{Equal contribution} \\
Superb AI\\
Seoul, South Korea\\
{\tt\small shkim@superb-ai.com}
\and
Kyeongryeol Go\footnotemark[1] \\
Superb AI\\
Seoul, South Korea\\
{\tt\small krgo@superb-ai.com}
}

\pgfplotsset{compat=1.18}
\begin{document}
\maketitle

\begin{abstract}
Fisheye cameras introduce significant distortion and pose unique challenges to object detection models trained on conventional datasets. In this work, we propose a data-centric pipeline that systematically improves detection performance by focusing on the key question of identifying the blind spots of the model. Through detailed error analysis, we identify critical edge-cases such as confusing class pairs, peripheral distortions, and underrepresented contexts. Then we directly address them through edge-case synthesis. We fine-tuned an image generative model and guided it with carefully crafted prompts to produce images that replicate real-world failure modes. These synthetic images are pseudo-labeled using a high-quality detector and integrated into training. Our approach results in consistent performance gains, highlighting how deeply understanding data and selectively fixing its weaknesses can be impactful in specialized domains like fisheye object detection.
\end{abstract}    
\begin{figure*}[t]
  \centering
  \includegraphics[width=0.95\textwidth]{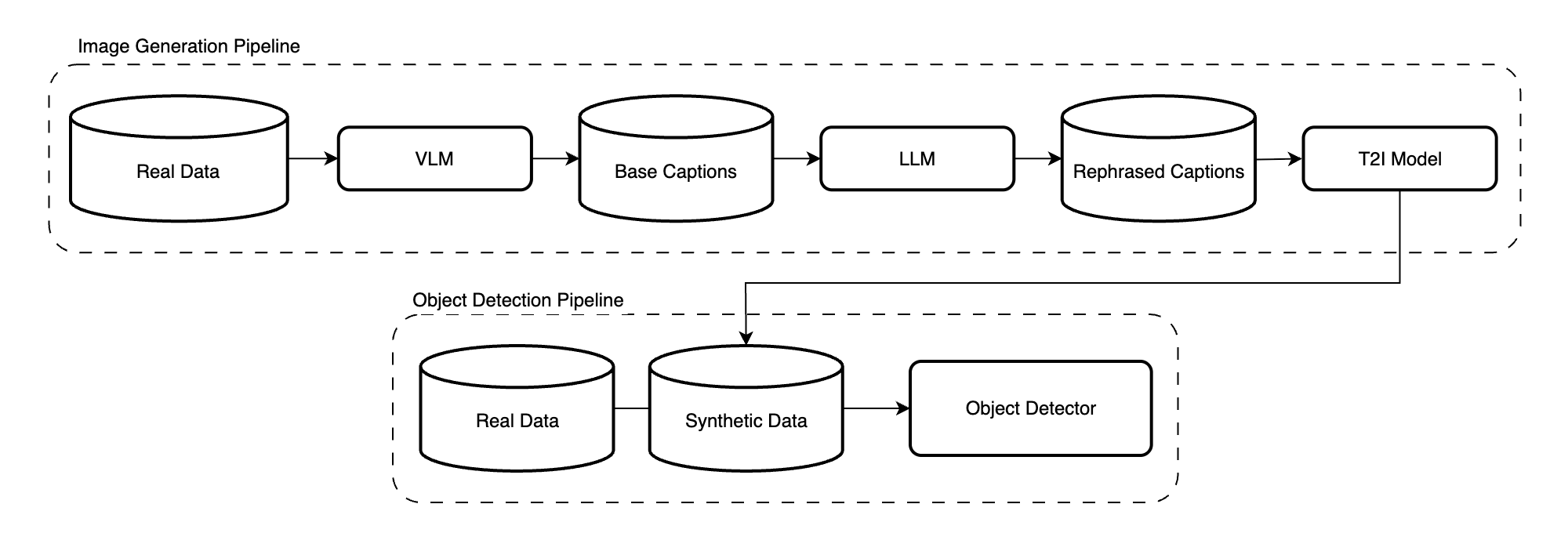}
  \caption{A simple diagram of our proposed pipeline. In addition to the dataset provided by the challenge, we incorporate external sources like VisDrone. Then, we leverage VLMs, LLMs, and text-to-image (T2I) models to generate a diverse set of synthetic data. All the datasets are annotated by the high-accuracy object detector, Co-DETR, as a pseudo-labeler, which contributes to robust performance improvement of a real-time object detector by knowledge distillation.}
  \label{fig:pipeline}
\end{figure*}

\section{Introduction}
\label{sec:intro-spb}

The increasing deployment of edge AI in traffic monitoring and smart mobility applications has brought renewed attention to fisheye cameras due to their wide-angle field of view (FoV). Compared to traditional perspective cameras, fisheye cameras can cover a significantly larger area, thereby reducing the number of cameras required in real-world deployments such as urban intersections and highways. This cost-effective and infrastructure-friendly characteristic makes them highly attractive for scalable traffic surveillance systems.

However, the use of fisheye cameras introduces new challenges. Their images suffer from strong radial distortions, making conventional image processing pipelines less effective. Correcting these distortions either requires computationally intensive undistortion techniques \cite{xu2023comprehensive} or dedicated model architectures that can directly process the distorted inputs.

To address this, the 9th AI City Challenge was organized to foster research in real-time object detection using fisheye cameras under diverse traffic conditions. The challenge adopts an evaluation metric that is the harmonic mean of F1-score and inference speed (FPS), encouraging participants to strike a balance between accuracy and computational efficiency. A minimum performance requirement of 10 FPS on a Jetson AGX Orin edge device is enforced to ensure practical deployability of the models.

The dataset provided in this challenge, FishEye8K \cite{gochoo2023fisheye8k} and FishEye1Keval, comprises fisheye images annotated for five traffic object classes: Bus, Bike, Car, Pedestrian, and Truck. The data reflects a wide range of traffic scenarios, including varying congestion levels, different road geometries such as intersections, and diverse lighting conditions across different times of day and viewing angles. The dataset is split into 5,288 training images, 2,712 validation images, and 1,000 test images.


An analysis of the dataset reveals significant imbalances across time of day. Notably, the Afternoon class dominates the training split, whereas Evening samples are absent. Night and Morning data are present but are limited to only one camera each, indicating limited scene diversity. Furthermore, the scale distribution is skewed such that most classes (especially Pedestrian and Bike) are heavily biased toward smaller scales. In addition, many object instances appear near the image boundaries where fisheye distortion is most prominent, further complicating detection.

This paper presents a comprehensive pipeline designed to enhance object detection performance for fisheye camera imagery, structured around a data-centric methodology. In Section~\ref{sec:related-spb}, we briefly review the overall research trends in real-time object detection, summarize the winning solutions from last year’s challenge, and discuss recent advances in synthetic data generation. Section~\ref{sec:method-spb} details each step of our pipeline: data collection, edge-case analysis, synthetic data generation, and data augmentation. Section~\ref{sec:exp-spb} presents implementation details, shows the incremental performance gains from each stage of data enhancement, conducts ablation studies on various training options, and reports the final results submitted to the challenge. Finally, Section~\ref{sec:conclu-spb} summarizes key insights and the effectiveness of our edge-case-focused, data-centric approach.
\section{Related work}
\label{sec:related-spb}

\subsection{Real-time object detection}
\label{sec:related-realtime-od}

Recent research in real-time object detection has largely followed two major directions: CNN-based approaches represented by the YOLO family \cite{redmon2016you, Jocher_Ultralytics_YOLO_2023} and transformer-based methods following the DETR paradigm \cite{carion2020end, zhao2024detrs, lv2024rt, peng2024d}.

In contrast to earlier two-stage detectors like Faster R-CNN \cite{girshick2015fast}, the YOLO (You Only Look Once) \cite{redmon2016you} model was the first to achieve real-time object detection by reformulating the task as a single regression problem. Instead of generating region proposals and refining them in multiple stages, YOLO performs detection in a single forward pass, enabling high-speed inference. Subsequent YOLO variants \cite{Jocher_Ultralytics_YOLO_2023} have continuously improved both efficiency and accuracy through architectural innovations and optimized training strategies, making them a dominant choice in real-time applications.

More recently, DETR (DEtection TRansformer) \cite{carion2020end} introduced a transformer-based paradigm that removes the need for hand-designed components such as anchor boxes and non-maximum suppression (NMS). Although DETR originally suffered from high computational cost and slow convergence, RT-DETR \cite{zhao2024detrs} addressed these limitations by introducing a hybrid encoder and minimal query selection strategy, achieving real-time performance for the first time among transformer-based detectors. Follow-up models \cite{lv2024rt, peng2024d} further improved detection accuracy and efficiency, positioning transformer-based detectors as strong alternatives to YOLO in the real-time detection landscape.

\subsection{Previous solutions}
\label{sec:related-previous}

In the 8th AI City challenge \cite{wang20248th}, there were no constraints on inference speed (FPS), allowing participating teams to prioritize accuracy by leveraging high-capacity models and extensive ensemble techniques.

The first-rank solution \cite{duong2024robust} investigated the use of additional datasets alongside the provided FishEye8K data to improve performance. After evaluating several candidates, they selected the VisDrone dataset \cite{du2019visdrone}, which showed strong complementary value. To better align the visual characteristics of VisDrone with the fisheye domain, they applied image–processing–based distortion techniques to create a synthetic fisheye version of VisDrone. They further applied pseudo-labeling using Co-DETR \cite{zong2023detrs} predictions on all training data to ensure label consistency. For the final ensemble, models such as YOLOv9 \cite{wang2024yolov9}, YOLOR-W6 \cite{wang2021you}, InternImage \cite{wang2023internimage}, and Co-DETR were trained on various data combinations and fused using Weighted Box Fusion (WBF) \cite{solovyev2021weighted}, resulting in a significant performance boost.

The second-rank solution \cite{shin2024road} focused on improving the detection of small objects, a known challenge in the FishEye dataset \cite{gochoo2023fisheye8k}. They adopted a tiling-based inference called SAHI \cite{akyon2022slicing} to better localize small-scale instances and reduce missed detections. To further reduce false positives, they added labels for non-target objects to help the model distinguish between relevant and irrelevant classes. In addition, they used super-resolution techniques using the StableSR model \cite{wang2024exploiting} to further upscale image resolution, improving detection quality and reducing false negatives. Like other top teams, they also employed model ensemble strategies to maximize final performance.

The third-rank solution \cite{pham2024improving} also tackled the false negative issue by incorporating background-only images from COCO \cite{lin2014microsoft} (images without objects) into training, helping the model better learn the distinction between background and object presence. To address the imbalance in day vs night images, they used CycleGAN \cite{zhu2017unpaired} to generate night-time images from day-time samples, effectively augmenting the dataset. Their final model was a YOLOR-D6 detector \cite{wang2021you} trained at three different resolutions (1280, 1536, and 1920), and combined using an ensemble. They also utilized test-time augmentation (TTA) \cite{shanmugam2021better} to enhance prediction robustness further.
 
However, in the 9th AI City Challenge, the final evaluation metric is defined as the harmonic mean of the F1 score and FPS measured on an edge device. This shift imposes practical constraints on computational efficiency, making it difficult to apply the accuracy-focused detectors and large-scale model ensembles used in previous solutions. In response, rather than relying on architectural complexity or ensemble methods, we focused on maximizing model performance with a data-centric approach under real-time constraints by maintaining a single lightweight detector.

\subsection{Synthetic data from generative models}
\label{sec:related-synthetic-data}

Recent advances in diffusion-based image generation models \cite{saharia2022photorealistic, rombach2022high, flux2024} have opened new opportunities for leveraging synthetic data to improve downstream tasks such as classification and object detection. \citet{he2022synthetic} was among the first to demonstrate that synthetic images generated by diffusion models can effectively support zero-shot and few-shot classification tasks, highlighting the potential of high-quality generative data in low-data regimes. Similarly, \citet{azizi2023synthetic} showed that fine-tuning the Imagen model \cite{saharia2022photorealistic} can enhance performance on large-scale classification benchmarks like ImageNet \cite{deng2009imagenet}. Beyond classification, there is growing interest in using diffusion-generated synthetic data to improve model robustness and generalization in more complex tasks such as object detection \cite{chen2023geodiffusion, tang2025aerogen} and semantic segmentation \cite{wu2023diffumask, nguyen2023dataset, wu2023datasetdm}.

While prior approaches for synthesizing object detection data \cite{li2023gligen, yang2023reco, zhou2024migc, wang2024instancediffusion} often rely on layout-guided generation to circumvent the need for manual annotations, such methods typically require extending pre-trained text-to-image models with additional modules specifically designed to process layout conditions. These approaches, though effective to a degree, face notable challenges, particularly in accurately rendering small or overlapping objects based on layout guidance \cite{chen2024region, zhou2025dreamrenderer}. In contrast, our method eliminates the need for explicit layout conditioning by relying solely on text prompts. This enables more lightweight and modular fine-tuning using parameter-efficient techniques, making it easier to adapt a wide range of pre-trained diffusion models without architectural modification.

Although our pipeline leverages a downstream detector to generate annotations, it remains more modular and tractable than alternative strategies that depend on attention maps \cite{wu2023diffumask} or latent embeddings \cite{wu2023datasetdm} from generative models for pseudo-labeling. Crucially, this modularity allows for independent improvements in both diffusion-based generation and object detection models to be seamlessly integrated, thereby facilitating the construction of higher-quality datasets.

Furthermore, while many existing methods employ filtering to ensure quality and diversity, our approach avoids such post-processing altogether. Instead, we rely on a targeted prompt engineering strategy informed by detailed error analysis, which enables the generation of edge-case scenarios. This design choice underscores the robustness of our synthesis pipeline and demonstrates its capacity to produce high-quality datasets without relying on additional curation steps.

\section{Methodology}
\label{sec:method-spb}

Rather than focusing on architectural modifications to the detection model, our approach to the challenge was fundamentally data-centric. We hypothesized that improving the training data distribution—especially in addressing underrepresented and challenging scenarios—would lead to more robust detection performance under the diverse and distorted conditions posed by fisheye cameras. To this end, we designed a step-by-step data-centric pipeline overcoming the unique challenges of fisheye object detection without relying on complex architectural changes. Figure \ref{fig:pipeline} provides a simplified illustration of the proposed method.

\subsection{Data collection}
\label{sec:method-data}

The FishEye8K dataset, captured entirely from fixed-position surveillance cameras, suffers from limited background diversity. Although it contains 8,000 images, only 14 unique camera locations are represented. This leads to rapid overfitting during training: when trained solely on FishEye8K, the model's performance saturates within 10 epochs, after which the validation loss begins to increase due to overfitting.

To enrich our dataset, we incorporated external sources containing relevant road traffic imagery. Specifically, we selected VisDrone \cite{du2019visdrone}, AAU RainSnow \cite{bahnsen2018rain}, UAVDT \cite{du2018unmanned}, and UA-DETRAC \cite{wen2020ua}, which were commonly captured from surveillance cameras or drones. Among these, VisDrone was fully utilized for training. However, the others were only partially used, as they contained numerous repetitive scenes with limited camera views. Therefore, these three datasets were only leveraged to extract base captions for synthetic data generation, which is described in Section~\ref{sec:method-syn}.

\begin{figure}[t]
    \centering
    \begin{subfigure}[t]{0.48\linewidth}
        \centering
        \includegraphics[width=\linewidth, height=4cm]{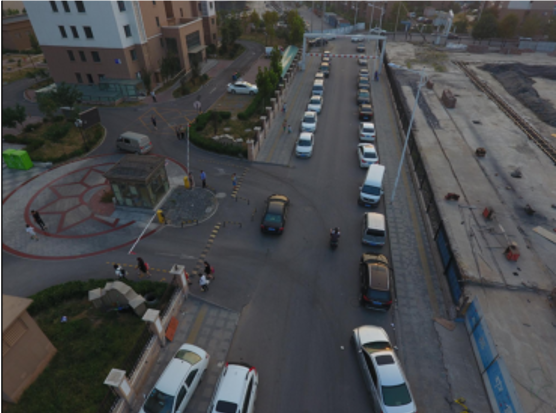}
        \caption{Non-fisheye image}
        \label{fig:non-fisheye}
    \end{subfigure}
    \hfill
    \begin{subfigure}[t]{0.48\linewidth}
        \centering
        \includegraphics[width=\linewidth, height=4cm]{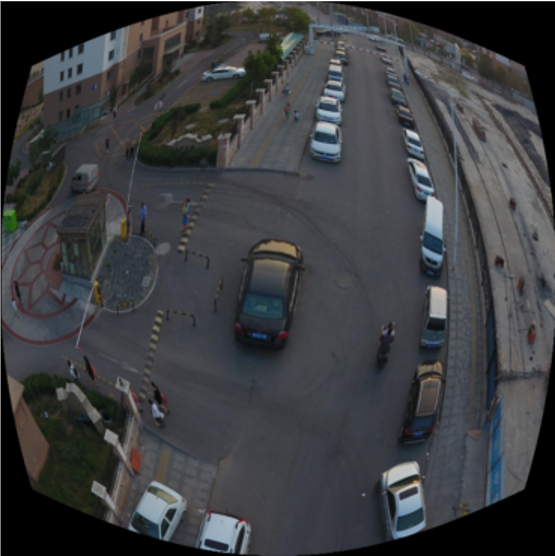}
        \caption{Fisheye-transformed image}
        \label{fig:fisheye-image}
    \end{subfigure}
    \caption{Comparison between a non-fisheye image in VisDrone and its fisheye-transformed counterpart. The transformation introduces radial distortion, mimicking fisheye camera effects.}
    \label{fig:fisheye}
\end{figure}

Although VisDrone introduced a greater variety of viewpoints and scene compositions into our training corpus, its images were not originally captured with a fisheye camera. To mitigate the risk of distribution shift, we implemented an augmentation pipeline that transforms regular images into fisheye-like images as visualized in \cref{fig:fisheye}. Since accurate calibration parameters were unavailable, we introduced randomness by sampling distortion coefficients and focal lengths within a plausible range.

\subsection{Edge-case analysis}
\label{sec:method-edge}

A comprehensive failure analysis was performed on the baseline detection model. This included visualizing false positives and false negatives on the validation set, examining confusing class pairs through the confusion matrix, and breaking down performance metrics by object class and contextual factors (\eg time of day, weather, location). Note that the contextual factors were extracted in advance by InternVL3-38B \cite{zhu2025internvl3} from the pre-defined set of candidates.

Upon analysis of the results, several edge-cases were identified as major contributors to performance degradation. One recurring issue was the difficulty in distinguishing visually similar categories such as Pedestrian vs Bike, Bus vs Truck, and Car vs Truck. Additionally, objects that were both small in size and located near the image boundaries—where fisheye distortion is most pronounced—were often misdetected. Another limitation was the underrepresentation of specific classes and contextual factors in the dataset.

\subsection{Synthetic data generation}
\label{sec:method-syn}

\begin{figure}[t]
    \centering
    \begin{subfigure}[t]{0.48\linewidth}
        \centering
        \includegraphics[width=\linewidth, height=4cm]{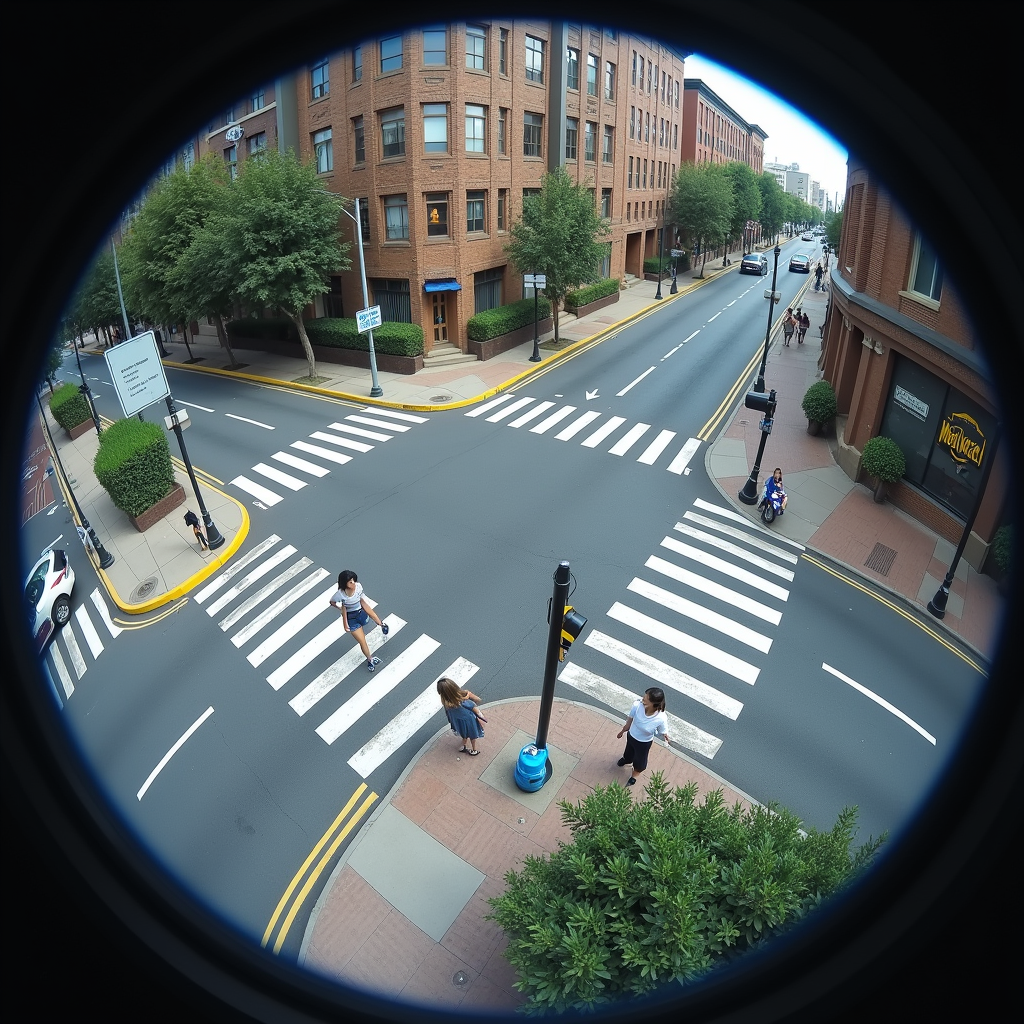}
        \caption{Image from pre-trained model}
        \label{fig:pretrain}
    \end{subfigure}
    \hfill
    \begin{subfigure}[t]{0.48\linewidth}
        \centering
        \includegraphics[width=\linewidth, height=4cm]{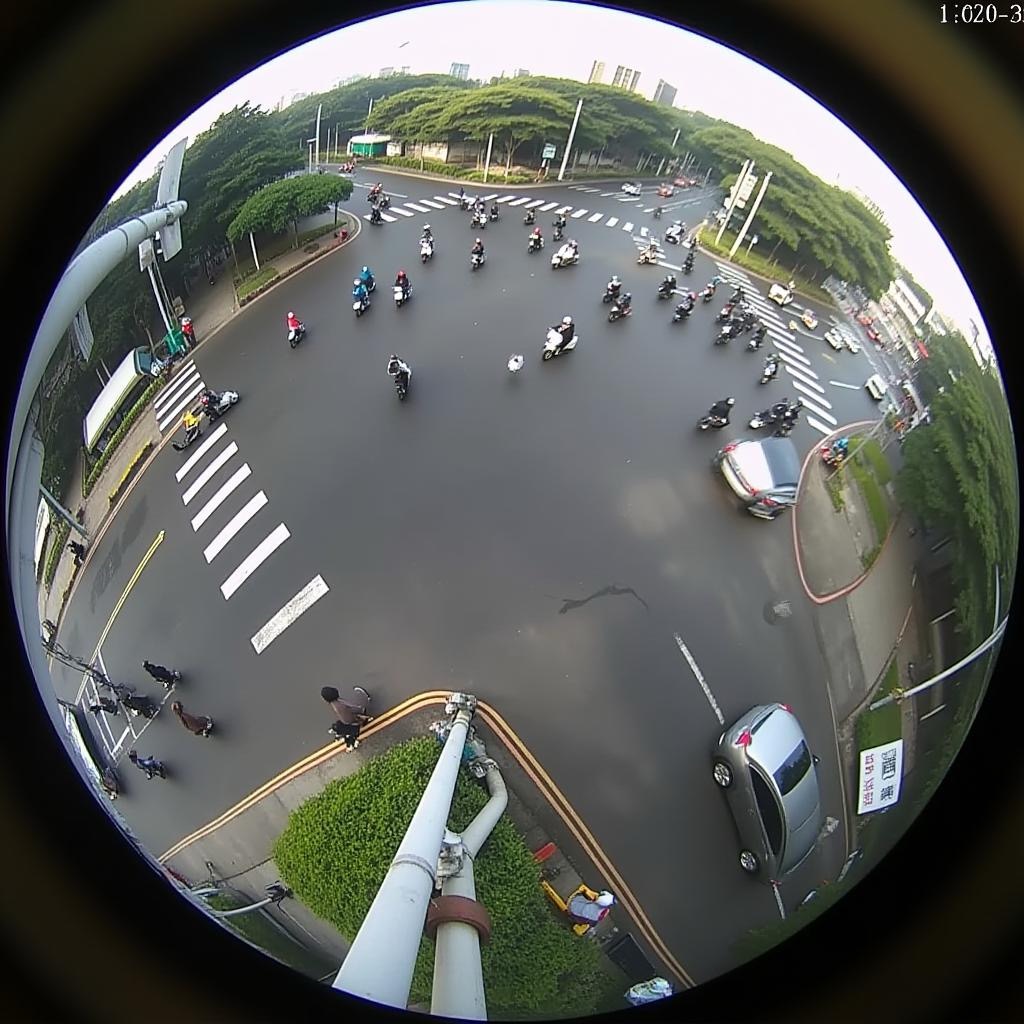}
        \caption{Image from fine-tuned model}
        \label{fig:finetune}
    \end{subfigure}
    \caption{(a): An image generated with a pre-trained generative model. (b): An image generated by prompting a fine-tuned generative model to insert many instances of Bike at the outer edges of the image, highlighting areas with severe fisheye distortion.}
    \label{fig:syn-example}
\end{figure}



We utilize image generation models to further enhance data diversity with images that closely resemble those captured with real fisheye lenses and that correspond to the edge-cases identified in \cref{sec:method-edge}. Among various generative model architectures, we choose a diffusion-based text-to-image model. These models are known for producing state-of-the-art image fidelity and offer strong controllability, allowing users to guide the generation process through natural language prompts. Particularly, we use FLUX.1-dev \cite{flux2024}, one of the highest-performing models currently available. However, images generated by FLUX.1-dev showed a mismatch in both distortion characteristics and overall scene composition compared to the FishEye8K domain (see \cref{fig:syn-example}). To reduce this domain discrepancy, we fine-tuned the model using the FishEye8K dataset. Details about configurations on fine-tuning FLUX.1-dev are described in \ref{sec:exp-implementation}.

In the generation phase, to ensure both diversity and relevance in the synthetic data, we carefully designed and refined prompts. First, to diversify backgrounds, we constructed a pool of base prompts derived from various surveillance datasets collected in \cref{sec:method-data}. Second, to specifically target edge-cases such as rare object classes, weather conditions, or peripheral occlusions, we used GPT-4.1-mini \cite{openai_gpt41_2024} to rephrase and enrich the base prompts according to the error patterns identified in \cref{sec:method-edge}. Details about the prompts for the rephrasing process are described in the supplementary material.

Because text-to-image models do not produce annotations, the resulting synthetic images lacked bounding box labels required for object detection training. To resolve this, we used Co-DETR \cite{zong2023detrs}, a high-accuracy object detector, to pseudo-label the synthetic data. This approach not only enables the effective use of generated images in supervised training but also allows for a form of knowledge distillation, where the real-time detector benefits from learning predictions made by a stronger model.


\subsection{Data augmentation}
\label{sec:method-aug}

We explored various data augmentation techniques. Building on the default configuration of Ultralytics \cite{Jocher_Ultralytics_YOLO_2023}, we evaluated augmentations that were disabled, such as vertical flip, copy-paste \cite{ghiasi2021simple}, and multi-scale resize. Note that to enable copy-paste augmentation, we generated mask annotations from bounding boxes using SAM2 \cite{ravi2024sam}. The final augmentation settings used for training the base model, along with related ablation results, are detailed in Section~\ref{sec:exp-ablation}.
\section{Experiments}
\label{sec:exp-spb}

We conducted experiments using YOLOv11s \cite{yolo11_ultralytics} as our base detection model. According to the challenge organizers, the TensorRT converted version of YOLOv11s achieves 62.33 FPS on a Jetson AGX Orin (64GB) device in 30W power mode with maximized clock frequencies. Since we use an input resolution of 1280 instead of the default 640, we estimate an effective inference speed of approximately 16 FPS—roughly one-quarter of the original speed.

\subsection{Implementation details}
\label{sec:exp-implementation}
For the image generative model, we utilize FLUX.1-dev \cite{flux2024}. We fine-tuned the model with the FishEye8K dataset, with captions obtained from InternVL3-38B \cite{zhu2025internvl3}. For the detailed prompts used with the VLM, please refer to the supplementary materials. To reduce memory and computational overhead during training while effectively retaining the knowledge of the pre-trained model, we employed LoKR \cite{yeh2023navigating}, a parameter-efficient fine-tuning method with rank of 16. The model was trained for 4,000 steps using the Prodigy optimizer. The initial learning rate was set to 1.0, with no warm-up steps, and a batch size of 16 was used.

The object detection model was trained for 100 epochs using SGD as the optimizer with an initial learning rate of 0.001. For data augmentation, we primarily followed the default settings provided by the Ultralytics framework. In addition, a fisheye transformation was probabilistically applied to samples from the VisDrone dataset to simulate distortion patterns observed in real-world fisheye imagery. Following the 1st place winner \cite{duong2024robust} in the 8th AI City Challenge, all training labels were unified under a single pseudo-labeling scheme generated by Co-DETR, regardless of whether the image was from real or synthetic data.

To fairly evaluate the effectiveness of each proposed component, all experiments were conducted using the validation split of the FishEye8K dataset for evaluation. The validation split was included in the training set only during the final model training stage, right before submission to the challenge. It is important to clarify that ground-truth annotations are used only when the validation split is used exclusively for evaluation to avoid any circular reasoning in model performance assessment. However, when the validation split is incorporated into the training data for the final model checkpoint for submission, it is annotated instead with pseudo labels generated by Co-DETR, ensuring consistency with the rest of the training set. Unless otherwise specified, all reported performance metrics in this paper are based on the F1 score at a confidence threshold of 0.5 and at IoU thresholds ranging from 0.50 to 0.95.

\subsection{Impact of incremental data addition}
\label{sec:exp-perf}


\begin{table}[t]
    \centering
    \begin{tabular}{l|c|c|c}
        \toprule
        Training dataset & Image count & mAP & F1 \\
        \midrule
        FishEye8K        & 5,288  & 0.3861   & 0.4058 \\
        +VisDrone        & 13,917 & 0.3896   & 0.4218 \\
        +Synthetic v1    & 22,546 & 0.4566   & 0.4682 \\
        +Synthetic v2    & 39,175 & 0.4686   & 0.4839 \\
        +Synthetic v3    & 49,134 & \textbf{0.4853}   & \textbf{0.4907} \\
        \midrule
        \rowcolor[HTML]{F5F5F5}
        Final submission & 51,846 & 0.4148   & 0.5119 \\
        \bottomrule
    \end{tabular}
    \caption{Performance comparison on overall mAP and F1 score.}
    \label{tab:ablation-dataset}
\end{table}

\begin{figure}[t]
    \centering
    \includegraphics[width=0.9\linewidth]{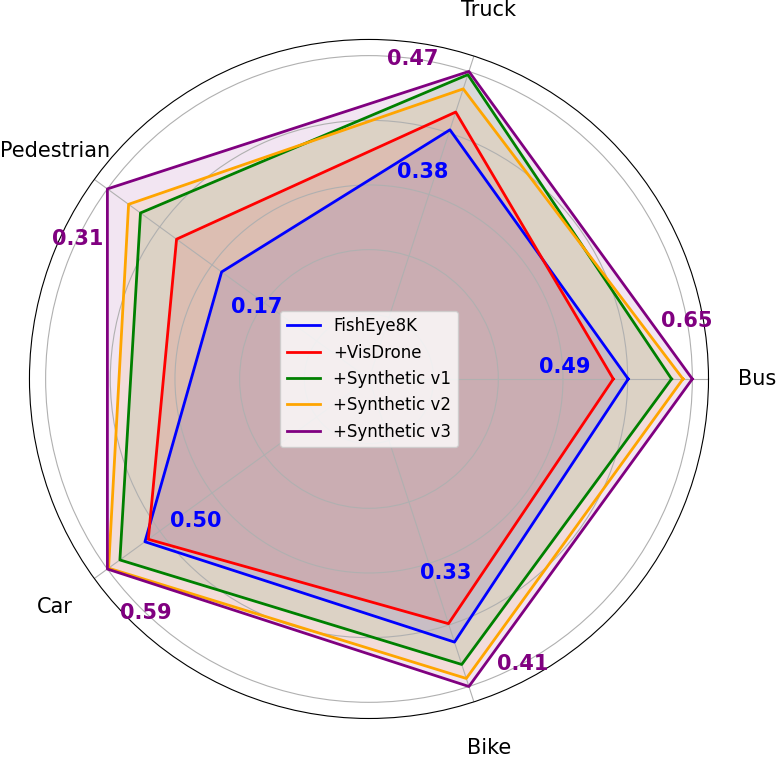}
    \caption{Radar plot illustrating class-wise AP across different training datasets. For visual clarity, class-wise AP values are annotated only for the datasets with the minimum and maximum volumes.}
    \label{fig:class-wise-ap}
\end{figure}

A key contribution of our work lies in demonstrating the effectiveness of a data-centric approach, particularly in addressing edge-cases through targeted dataset expansion. To validate this, we conducted a series of experiments where training data progressively expanded, and the model’s performance was evaluated at each stage. A clear trend of gradual improvement was observed as additional real and synthetic data were incorporated.

We began with the baseline setting using only the FishEye8K dataset. Next, we added the VisDrone dataset, which introduces greater viewpoint and scene diversity, albeit captured with non-fisheye cameras. Building upon this, we introduced \textbf{Synthetic v1}, a set of generated images conditioned on VisDrone captions, designed to enhance overall diversity. Following that, we added \textbf{Synthetic v2}, synthesized using captions from both FishEye8K and VisDrone, with prompts carefully crafted to reflect general edge-case characteristics identified in our error analysis. Finally, \textbf{Synthetic v3} was introduced, targeting specific edge-cases by incorporating captions from additional datasets—AAU RainSnow, UA-DETRAC, and UAVDT—and refining prompts to emphasize under-represented categories and contexts. Please refer to the supplementary material for details on the prompts that are used for each stage of synthetic data generation.

\begin{figure*}[t]
    \centering
    \begin{subfigure}[b]{0.3\textwidth}
        \centering
        \includegraphics[width=\linewidth]{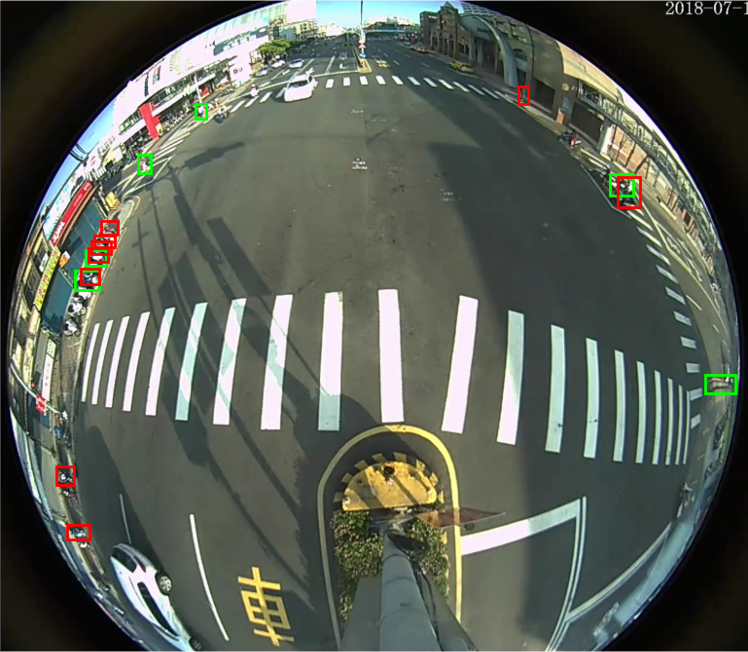}
        \caption{+Synthetic v1}
    \end{subfigure}
    \hfill
    \begin{subfigure}[b]{0.3\textwidth}
        \centering
        \includegraphics[width=\linewidth]{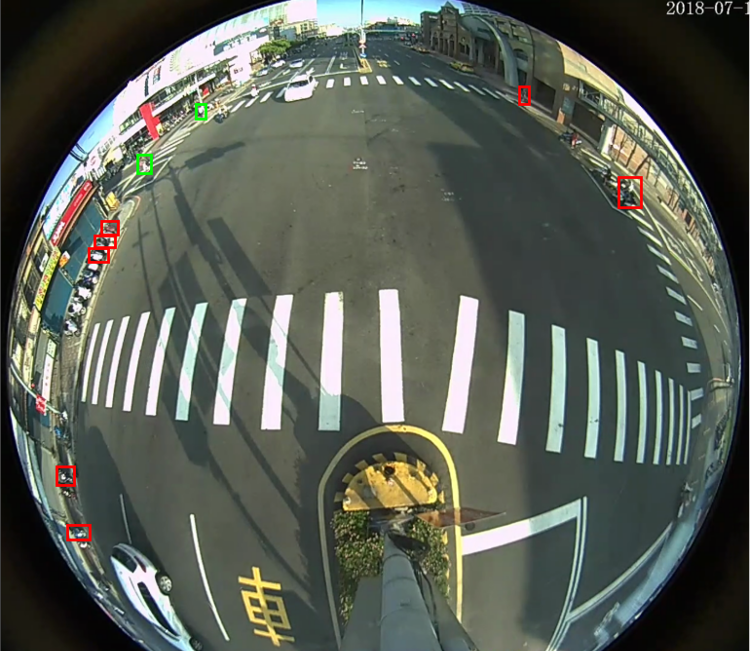}
        \caption{+Synthetic v2}
    \end{subfigure}
    \hfill
    \begin{subfigure}[b]{0.3\textwidth}
        \centering
        \includegraphics[width=\linewidth]{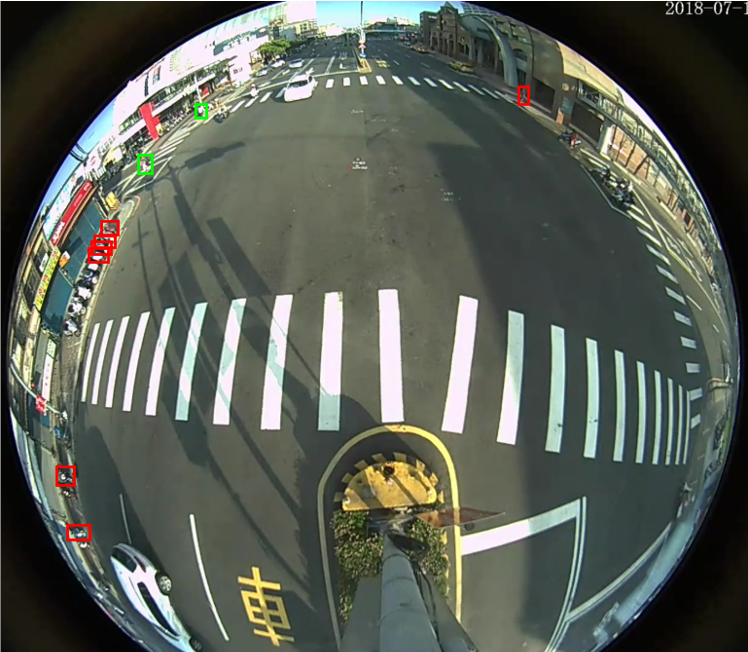}
        \caption{+Synthetic v3}
    \end{subfigure}
    \caption{Qualitative comparison of false positives and false negatives as more synthetic data is added. False positives are marked in \textcolor{lime}{lime}, and false negatives in \textcolor{red}{red}. Best viewed in color.}
    \label{fig:fpfn}
\end{figure*}

Incremental addition of the training data led to consistent performance improvements, as evidenced by \cref{tab:ablation-dataset}. In particular, the introduction of Synthetic v1 resulted in significant gains in both mAP and F1 score. As illustrated in \cref{fig:class-wise-ap}, the inclusion of Synthetic v2 and v3 further enhanced performance, with the notable percentage improvements observed in previously under-performing classes such as Pedestrian. These findings underscore the efficacy of targeted data synthesis in addressing rare or challenging categories. Furthermore, \cref{fig:fpfn} qualitatively demonstrates how false-positive and false-negative predictions are corrected as more synthetic data is incorporated.

For final submission, we applied class-wise optimal confidence thresholds to filter predictions, aiming to maximize the F1 score on the test set. These thresholds were determined based on Co-DETR pseudo labels for the test split, which served as a reference to approximate the true distribution in the absence of ground-truth annotation. Note that this was the only case where we utilized the test split throughout the whole experiment. The test split was never used for model training or for tuning training-related hyperparameters to preserve its integrity and prevent any potential issues related to fairness, data leakage, or biased evaluation.

When submitted to the challenge evaluation server, our final model, trained with the complete dataset including all synthetic versions and also the validation split of the FishEye8K dataset, achieved the F1 score of 0.5119. This result highlights the impact of strategic, edge-case-aware data synthesis and augmentation in real-world performance without architectural modifications.

\subsection{Ablation study}
\label{sec:exp-ablation}

To understand the impact of individual design choices in our pipeline, we conducted extensive ablation studies across two key components: label strategy and data augmentation. These experiments aimed to isolate the contribution of each factor to the overall performance.

\begin{table}[t]
    \centering
    \begin{tabular}{l|c}
        \toprule
        Label strategy & F1 \\
        \midrule
        Ground Truth & 0.4660 \\
        Pseudo Label & \textbf{0.4907} \\
        Pseudo + Background Label & 0.4796 \\
        \bottomrule
    \end{tabular}
    \caption{Comparison of different label strategies.}
    \label{tab:ablation-lab}
\end{table}

\paragraph{Label strategy} We began by assessing the impact of label type by comparing models trained on ground-truth annotations with those trained on pseudo-labels produced by Co-DETR. \cref{tab:ablation-lab} reported the experiments conducted with all the available training datasets and revealed that pseudo-labels consistently outperformed the original annotations, indicating potential label noise in the dataset and underscoring the reliability of our pseudo-labeling approach. Motivated by the methodology of the 3rd place winner \cite{pham2024improving} in the 8th AI City Challenge, we further incorporated background annotations to mitigate false positives. To achieve this, we employed an open-set object detector, Grounding-DINO \cite{liu2024grounding}, in a zero-shot setting to detect instances across the five target categories. Detections with low intersection-over-union (IoU) compared to the pseudo-labels generated by Co-DETR were designated as background annotations, as they likely corresponded to potential false positives. However, empirically, background annotations appear to hinder performance rather than improve it.

\begin{table}[t]
    \centering
    \begin{subtable}[t]{0.48\linewidth}
        \begin{minipage}[t]{\linewidth}
            \centering
            \begin{tabular}{l|c}
                \toprule
                Augmentation & F1 \\
                \midrule
                vertical flip & 0.3154 \\
                copy-paste & 0.3155 \\
                MSR & 0.3234 \\
                \midrule
                default & \textbf{0.3256} \\
                \bottomrule
            \end{tabular}
            \caption{Standard augmentations}
            \label{tab:aug_std}
        \end{minipage}
    \end{subtable}
    \hfill
    \begin{subtable}[t]{0.48\linewidth}
        \begin{minipage}[t]{\linewidth}
            \centering
            \begin{tabular}{l|c}
                \toprule
                Augmentation & F1        \\
                \midrule
                non-fisheye  & 0.4791    \\
                fisheye      & \textbf{0.4907} \\
                \bottomrule
            \end{tabular}
            \caption{Fisheye augmentation}
            \label{tab:aug_fisheye}
        \end{minipage}
    \end{subtable}
    \caption{Ablation study on data augmentation techniques. }
    \label{tab:ablation-augment}
\end{table}


\paragraph{Data augmentation} In \cref{tab:aug_std}, we evaluated various augmentation techniques to determine their individual contributions. Using the 640 resolution FishEye8K dataset, we tested augmentations such as vertical flip, copy-paste, and Multi-Scale Resize (MSR). However, none of these augmentations led to meaningful performance improvements when tested individually. We also assessed the effect of applying fisheye transformations as an augmentation in \cref{tab:aug_fisheye}. Using all available training datasets, we compared models trained with and without the fisheye transform applied to the VisDrone dataset. As a result, the model augmented with the fisheye transformation showed superior performance.




\section{Conclusion}
\label{sec:conclu-spb}

In this work, we presented a data-centric approach to improving real-time object detection performance in fisheye camera imagery. Rather than relying on model-level innovations, we systematically tackled data-related weaknesses mainly through edge-case analysis and synthetic data generation. Specifically, we synthesized fisheye-style images tailored to the failure modes of a baseline detector with a fine-tuned text-to-image generative model. Our findings demonstrate that edge-case synthesis, when guided by thorough data analysis and precise prompt engineering, can meaningfully enhance model performance in underrepresented scenarios.

Building on this foundation, several promising directions remain open. A more nuanced treatment of background labels, along with the integration of additional training strategies aimed at mitigating false positives and false negatives, could further refine detection accuracy. It is also important to characterize how performance scales with the volume of generated data, particularly identifying the point at which additional samples yield diminishing returns. Furthermore, while our study relied on a single generative model (Flux.1-dev), leveraging a broader range of text-to-image generators may improve the diversity and realism of synthetic samples, and potentially delay the onset of performance saturation. These avenues constitute a natural next step toward fully harnessing synthetic data for robust fisheye perception.
{
    \small
    \bibliographystyle{ieeenat_fullname}
    \bibliography{main}
}

\clearpage


\section*{Supplementary material}

# Supplementary Material for Edge-case Synthesis for Fisheye Object Detection

Seunghyeon Kim*
Superb AI
Seoul, South Korea
shkim@superb-ai.com

Kyeongryeol Go*
Superb AI
Seoul, South Korea
krgo@superb-ai.com

## 1. Prompt engineering for base captions

In this section, we provide the prompt used with the InternVL3-38B model to obtain the base captions. We specifically guided the model to ensure that the base captions included the key objects targeted in the fisheye dataset. The captions obtained in this way were used both for training the generative model and as base captions for subsequent rephrasing.

> You are an expert in generating high-quality image captions. Please analyze the provided fish-eye image in detail.
>
> Please adhere to the following format for the caption:
> - Start with "A photo of".
> - Limit the total length to 40-50 words.
> - Focus on bus, bike, car, pedestrian and truck.
> - Describe time of day, weather and location.
> - Focus on the scene inside the fish-eye lens.
> - Use grammatically correct and clear sentences.

## 2. Prompt engineering for rephrased captions

In this section, we provide the detailed prompts used in our pipeline. These prompts were given to GPT-4.1-mini to rewrite base captions that are originally extracted from real data so that they better describe the types of images we aimed to generate.

### 2.1. Synthetic v1

In Synthetic v1, the primary goal is to increase overall data diversity. To achieve this, we retain most of the base captions derived from the real dataset, but modify them slightly to describe scenes as if they were captured from a fisheye view.

> You are an expert in generating high-quality image captions. Please rewrite the caption I give you.
>
> Please adhere to the following format for the caption:
> - Rewrite the caption to fish-eye view.
> - Start with "A photo of".
> - Limit the total length to 40–50 words.
> - Use grammatically correct and clear sentences.

### 2.2. Synthetic v2

In Synthetic v2, we introduced more fine-grained prompting strategies to guide caption generation. The instructions explicitly encourage highlighting key traffic entities—Bus, Bike, Car, Pedestrian, and Truck—particularly in the peripherally distorted regions where small-scale objects are most affected by fisheye lens geometry. To reduce ambiguity, we provided precise

---

*Equal contribution

object definitions and disambiguation heuristics between visually similar categories. Additionally, by embedding realistic traffic scene layouts and diverse environmental conditions (e.g., lighting, intersection types, and traffic flow), along with a slight emphasis on semantically important yet underrepresented classes such as Truck and Pedestrian, we fostered the creation of captions that are both varied and semantically grounded. This approach ultimately facilitates the training of object detection models that are more resilient to fisheye-induced distortions and better suited for real-world deployment in complex urban environments.

> You are an expert at visually-grounded image caption rewriting. Your task is to rewrite image captions taken with a fisheye camera so that:
> - The objects **Bus, Bike, Car, Pedestrian, and Truck** are **small in scale** and located near the **edges of the image**, where **fisheye distortion** is strong.
> - Object placement and interaction should be **plausible** within real-world traffic scenes.
>
> Apply the following object descriptions:
> - **Bus**: large public passenger vehicles
> - **Truck**: heavy-duty vehicles like dump trucks or semi-trailers
> - **Car**: compact vehicles such as sedans, SUVs, or vans
> - **Bike**: bicycles, motorcycles, or scooters, either parked or with riders
> - **Pedestrian**: visible people walking, standing, or crossing
>
> Avoid visual ambiguity between:
> - **Bus vs Truck** → contrast size, function, silhouette
> - **Car vs Truck** → emphasize bulk and height differences
> - **Pedestrian vs Bike** → distinguish by motion, vehicle presence, posture
>
> Ensure variation across scene conditions:
> - **Camera angles**: side-view or front-view
> - **Intersection types**: T-junctions, Y-junctions, cross-intersections, mid-blocks, pedestrian crossings, or irregular layouts
> - **Lighting**: morning, afternoon, evening, or night
> - **Traffic flow**: free-flowing, steady, or busy
>
> When choosing scene elements, slightly **favor** the following (but still maintain diversity overall):
> - **Categories prominently shown**: **Pedestrian** and **Truck**
> - **Time of day**: **Day, Afternoon, Night, or general Daytime**
> - **Weather**: **Clear**
> - **Location type**: **Urban areas such as City streets**
>
> Preserve the core content of the original caption, but rewrite it to reflect the above constraints. If none of the specified categories are present, you may subtly introduce one or more at the distorted outer edges of the image. Always maintain natural, fluent language, and don't make the added objects the main focus unless already emphasized.

### 2.3. Synthetic v3

Finally, in Synthetic v3, we revised the scene descriptions to ensure that objects of specific target classes appear frequently in particular backgrounds. The focus is placed on targeted generation over sheer diversity, aiming to maximize informational effectiveness on a per-image basis. Below is an example prompt used to generate a caption where Bike frequently appears in a Y-junction setting.

> You are an expert in generating high-quality image captions. Please rewrite the caption I give you.
>
> Please adhere to the following format for the caption:
> - Rewrite the caption to fish-eye view.
> - Start with *"A photo of"*.
> - Limit the total length to 40–50 words.
> - Use grammatically correct and clear sentences.
> - Add many **Bike** in the scene (*Bike*: bicycles, motorcycles, or scooters, either parked or with riders).
> - Change location to **Y-junctions**.

## 3. Fine-tuned synthetic image quality

To better illustrate the effectiveness of our data generation pipeline, we provide qualitative examples of synthetic images generated before and after fine-tuning the diffusion model. Figure 1 presents five representative examples, where each column compares images generated from the base pre-trained model (top) and our fine-tuned model (bottom).

As shown, the images synthesized by the pre-trained model often diverge from the visual characteristics of the target domain. In contrast, the fine-tuned model generates images that are not only more visually aligned with the FishEye8K dataset but also successfully reflect the intended object categories with appropriate visual context. Notably, the fine-tuned images demonstrate diversity in terms of background elements such as lightning, suggesting that the fine-tuning generalizes well without overfitting.

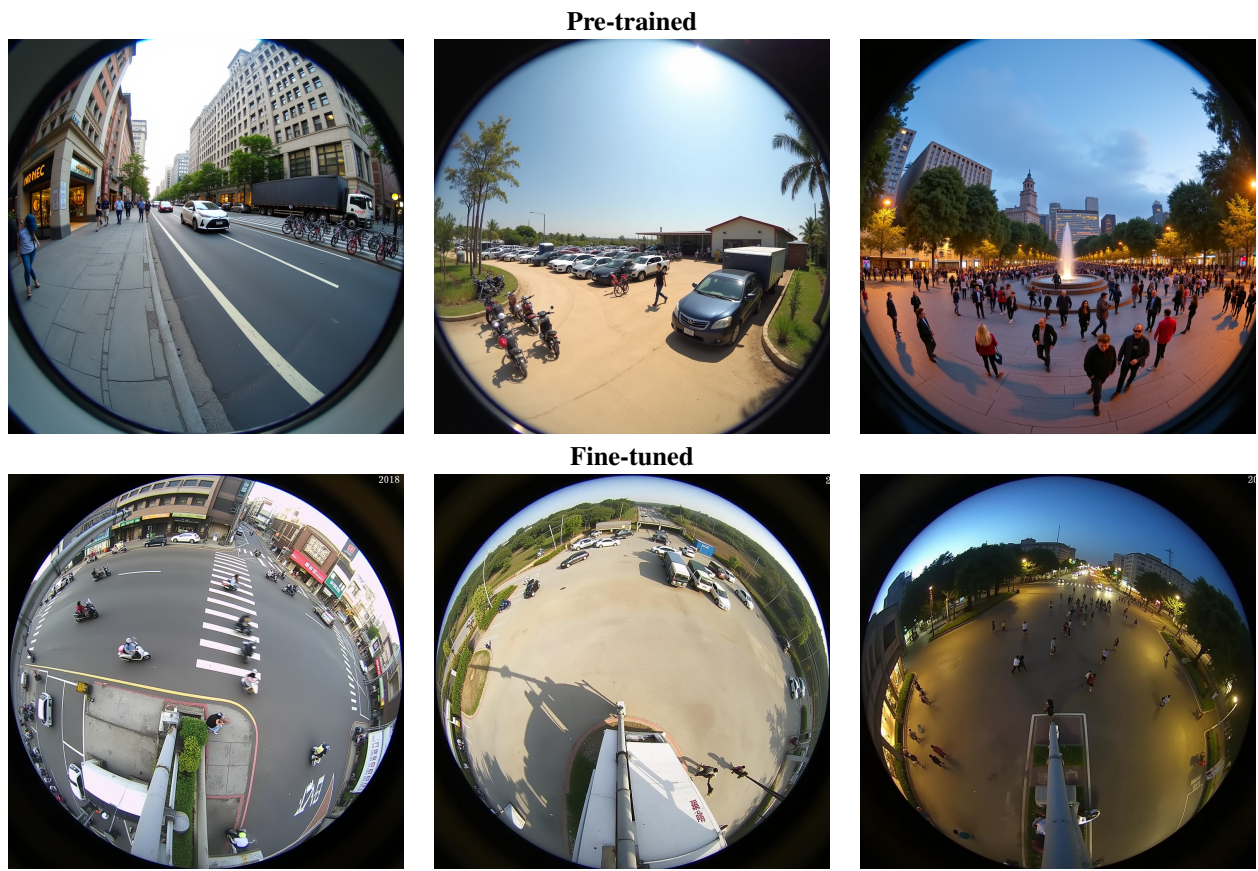

Figure 1. Qualitative comparison of synthetic images generated by the pre-trained diffusion model (top) and the fine-tuned model (bottom). Our fine-tuned model produces images more consistent with the target domain, while also preserving background diversity without overfitting.

## 4. Effect of synthetic data on prediction quality

To further demonstrate the effectiveness of our synthetic data in improving detection performance, we visualize model predictions on three different validation images under incremental incorporation of synthetic datasets (v1, v2, v3). Figure 2 shows that as the synthetic data improves, both false positives and false negatives are gradually reduced, indicating more accurate and confident detection.

This trend is consistent across a wide range of validation images, suggesting that our synthesized data not only improves the model's overall performance but also enhances its generalization to unseen cases. Notably, many of the false detections present in earlier versions of synthetic data are corrected as the generation process becomes more tailored and diverse.

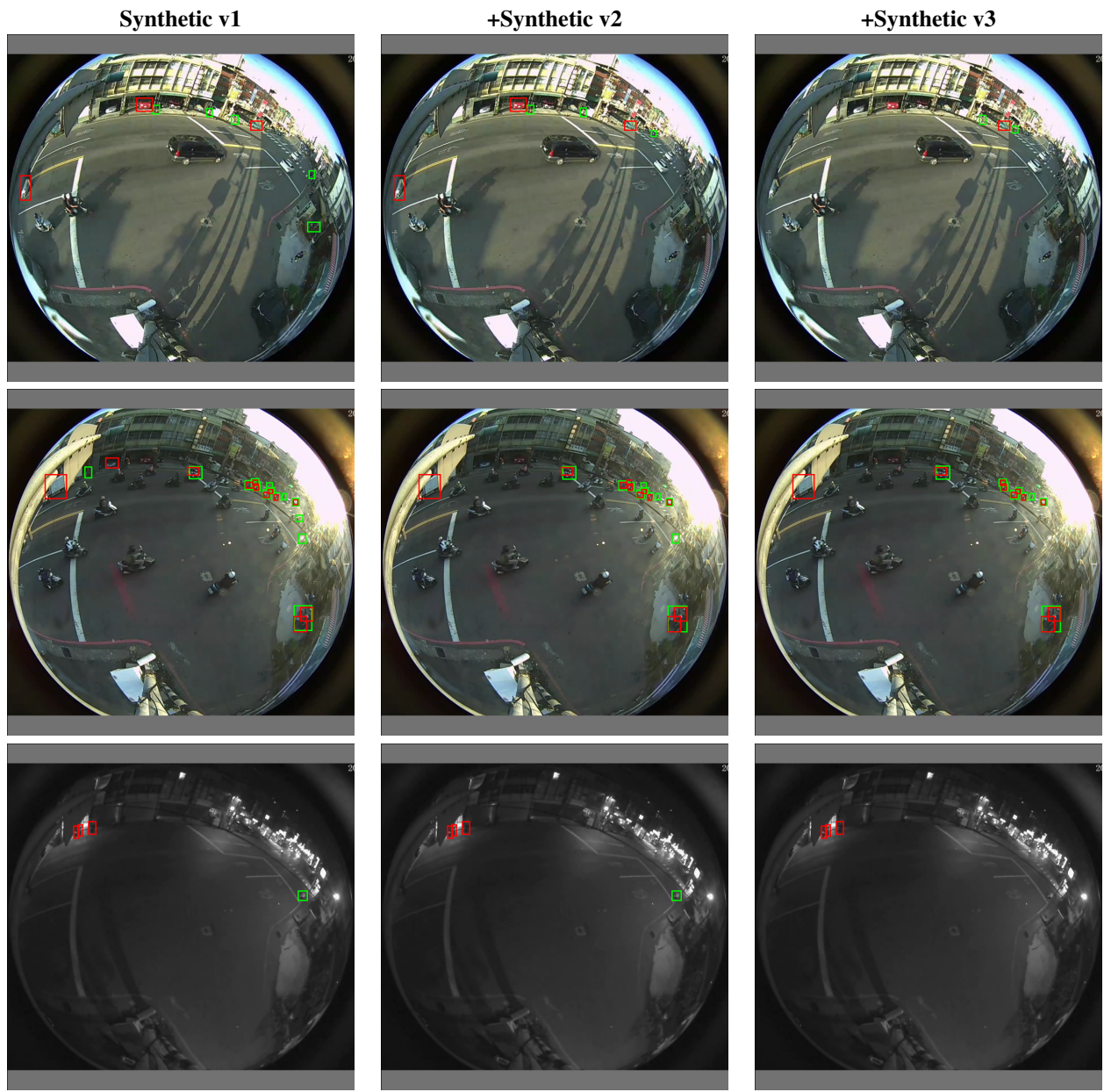

Figure 2. Comparison of model predictions on validation images using synthetic data v1 (left), v2 (center), and v3 (right). False positives are marked in lime, and false negatives in red. Best viewed in color. Across different images, false positives and false negatives decrease as higher-quality synthetic data is incorporated, indicating better generalization and robustness.



\end{document}